\documentclass[letterpaper, 10 pt, journal, twoside]{IEEEtran}

\usepackage[T1]{fontenc}

\usepackage{graphicx}      
\usepackage{pict2e}

\usepackage{graphics, dblfloatfix} 
\usepackage{epsfig} 
\usepackage{amsmath} 
\usepackage{amssymb} 
\usepackage{amsthm}
\usepackage{epstopdf, cases}

\usepackage{siunitx} 
\usepackage{subfig}			
\usepackage[section]{placeins}	
\usepackage{textcomp}			
\usepackage{booktabs}	
\usepackage{multirow}
\usepackage{tikz}
\usetikzlibrary{matrix}
\usepackage{nicefrac}       

\usepackage{listings}
\usepackage[numbered, framed]{matlab-prettifier}

\usepackage{scalerel}
\usepackage{times}

\usepackage[most]{tcolorbox}
\usepackage{enumitem}

\newtheorem{remark}{Remark} 
\DeclareMathOperator*{\argmin}{argmin}

\definecolor{rosso}{rgb}{1, 0, 0}
\definecolor{nero}{rgb}{0, 0, 0}
\definecolor{grigio}{rgb}{0.6, 0.6, 0.6}
\definecolor{blu}{rgb}{0, 0, 1}

\newcommand{\rev}[1]{\textcolor{nero}{#1}}
\newcommand{\revout}[1]{}

\newtheorem{problem}{Problem}

\colorlet{helpful}{lime!70}
\colorlet{harmful}{red!30}
\colorlet{internal}{yellow!20}
\colorlet{external}{cyan!30}
\colorlet{S}{helpful!50!internal}
\colorlet{W}{harmful!50!internal}
\colorlet{O}{helpful!50!external}
\colorlet{T}{harmful!50!external}
\colorlet{SO}{S!50!O}
\colorlet{WO}{W!50!O}
\colorlet{ST}{S!50!T}
\colorlet{WT}{W!50!T}

\tcbset{
    mybox/.style 2 args={%
        enhanced, 
        sharp corners, notitle, 
        before skip=6pt, after skip=6pt, 
        watermark text=#1, frame hidden, 
        colback=#2, 
    }, 
    mybox/.default={mybox={A}{white}}, 
}

    {\end{itemize}}

\begin{document}
\title{Safety-aware time-optimal motion planning with uncertain human state estimation}


\markboth{IEEE ROBOTICS AND AUTOMATION LETTERS. PREPRINT VERSION. ACCEPTED SEPTEMBER, 2022}
{Faroni \MakeLowercase{\textit{et al.}}: Safety-aware time-optimal motion planning with uncertain human state estimation} 

\author{\large Marco Faroni$^1$, Manuel Beschi$^{1, 2}$, Nicola Pedrocchi$^1$ %

\thanks{Manuscript received: May 13, 2022; Revised:
August 10, 2022; Accepted: September 12, 2022.

This paper was recommended for publication by
Editor Gentiane Venture upon evaluation of the Associate Editor and Reviewers’ comments.

This work is partially supported by ShareWork project (H2020, European Commission -- G.A. 820807).} 
\thanks{$^{1}$Marco Faroni, Manuel Beschi, and Nicola Pedrocchi are with Istituto di Sistemi e Tecnologie Industriali Intelligenti per il Manifatturiero Avanzato, Consiglio Nazionale delle Ricerche -- Milano, Italy
        {\tt\footnotesize \{marco.faroni, nicola.pedrocchi\}@stiima.cnr.it}}%
\thanks{$^{2}$Manuel Beschi is with Dipartimento di Ingegneria Meccanica e Industriale, 
University of Brescia -- Brescia, Italy
    {\tt\footnotesize manuel.beschi@unibs.it}}%
\thanks{Digital Object Identifier (DOI): see top of this page.}
}


\maketitle

\begin{abstract}
Human awareness in robot motion planning is crucial for seamless interaction with humans.
Many existing techniques slow down, stop, or change the robot's trajectory locally to avoid collisions with humans.
Although using the information on the human's state in the path planning phase could reduce future interference with the human's movements and make safety stops less frequent, such an approach is less widespread.
This paper proposes a novel approach to embedding a human model in the robot's path planner.
The method explicitly addresses the problem of minimizing the path execution time, including slowdowns and stops owed to the proximity of humans.
For this purpose, it converts safety speed limits into configuration-space cost functions that drive the path's optimization.
The costmap can be updated based on the observed or predicted state of the human. 
The method can handle deterministic and probabilistic representations of the human state and is independent of the prediction algorithm.
Numerical and experimental results on an industrial collaborative cell demonstrate that the proposed approach consistently reduces the robot's execution time and avoids unnecessary safety speed reductions.
\end{abstract}



\section{Introduction}
\label{sec:intro}
\IEEEPARstart{C}{ollaboration} between humans and robots requires a paradigm change in robots' perception, reasoning, and action.
From a planning point of view, the robot should be aware of the human's behavior and act accordingly.
Planning frameworks have embodied human-awareness at different levels, spanning ontologies \cite{Umbrico:ontology}, task planning \cite{Makris:task-planning-hrc}, scheduling \cite{Makris:scheduling-hrc}, motion planning \cite{Lasota:hamp}, and control \cite{haw-control}.
In motion planning, the robot should consider the human's state to avoid collisions, enhance human comfort, or improve efficiency.

Two main approaches are in the literature.
A first approach can be referred to as \emph{reactive}, as it modifies a given trajectory, during its execution, according to the current state of the human.
Existing works do so by adjusting the path on the fly \cite{SAVERIANO201796, Tonola_ROMAN2021} or, more frequently, by modifying the robot speed along a given path \cite{Zanchettin:safety, RCIM:implementing-ssm, RCIM:dynamic-ssm, Palleschi:safe-time-optimal}.

The second approach aims to plan trajectories thought to anticipate the humans' actions in such a way as to assist or avoid them.
For example, based on a prediction model, the planner may seek to avoid regions where the operator is expected to be.
This category can be referred to as \emph{proactive} \cite{Zanchettin:proactive-motion-planning}.
Notice that the two approaches can co-exist: a \emph{proactive} layer finds a human-aware trajectory, and a \emph{reactive} one adjusts it at run-time, considering unexpected events. 

This paper proposes a \emph{proactive} human-aware path planner that defines a cost function to minimize the trajectory execution time by including the expected safety slowdown owed to human-robot proximity.

\subsection{Related works}

Sisbot et al. \cite{Alami:hamp} and Mainprice et al. \cite{Mainprice:hamp} proposed a human-aware cost function that considers the human-robot distance, the human field of view, and human comfort.
The resulting motion planning problem is solved by using Transition-RRT \cite{Simeon-T-RRT}, a sampling-based algorithm for custom cost functions.
Hayne et al. \cite{Berenson:human-robot-planning} minimizes a cost function to avoid regions of the workspace previously occupied by the human and to favor repeatability of the robot motion (\emph{i.e.}, the cost function penalizes trajectories far from the previous ones).
Similarly, Zhao and Pan  \cite{Zhao:considering-human-behavior} use STOMP \cite{STOMP:2011} to minimize a cost function that penalizes previously occupied regions and to maximize the human-robot distance.
Casalino et al. \cite{Zanchettin:proactive-motion-planning} deform the trajectory according to a repulsion field associated with checkpoints on the human skeleton.
Tarbouriech and Suleiman \cite{Tarbouriech:bi-objective} propose a sampling-based algorithm that biases the search towards regions far from the current human state.
Javdani et al. \cite{Pellegrinelli:POMDP-hindsight-optimization} and Kanazawa et al. \cite{Kanazawa:TRO2019} pose the problem of reducing the execution time of robot actions by avoiding idle times in handover tasks.
For example, \cite{Kanazawa:TRO2019} uses model predictive control to reach a target state with a given optimized schedule, despite it does not consider collisions with the environment.
Other works focused on the evaluation of human factors when using human-aware motion planning, showing an enhanced work fluency and operators' satisfaction \cite{Lasota:hamp,Faroni_CPHS2020}.

A significant issue in human-aware methods is how to define a meaningful cost function.
All the methods above realize an arbitrary compromise between some desired cost terms.
For example, \cite{Tarbouriech:bi-objective, Wang:optimal-motion-planning} find a trade-off between the minimization of the path length and the maximization of the human-robot distance, while \cite{Kanazawa:TRO2019} minimizes a trade-off of the final error, joint bound satisfaction, and human-robot collision probability.
However, this approach does not model the delays caused by safety stops and slowdowns when the human is close to the robot. Moreover, the solution strongly depends on the weight tuning. Notably, the prediction of the human's motion comes with a significant uncertainty owed to the accuracy of human tracking systems and the intrinsic unpredictability of human behavior.
Therefore, a human-aware planner should consider the probabilistic representation of the operator's motion.

Finally, it is desirable that general-purpose motion planning algorithms can integrate human-aware motion planning methods.
\cite{Alami:hamp} and \cite{Mainprice:hamp} go in this direction by defining a cost function dependent on the robot configuration and the human's state and using it in a sampling-based path planner.

\subsection{Contribution}
\label{subsec:contribution}

This paper addresses the problem of planning paths with minimum expected execution time.
Differently from existing works, the paper proposes a cost function that approximates the path execution time, including safety speed reductions caused by human proximity. 
To do so, it converts the expected limitations of the robot speed -- as defined in ISO-TS/15066 -- into a delay in the nominal execution time (Section \ref{sec:method}). 
By solving a path planning problem with this cost function, it is possible to minimize the expected execution time, instead of an arbitrary trade-off of several costs. 
In this regard, the method can be used with any configuration space-based motion planners.
The method considers both deterministic and probabilistic representations of the the human's state  (Section \ref{subsec:deterministic-cost-function} and \ref{subsec:probabilistic-cost-function}) and is independent of the model used to predict the human state. 
It defines a voxelized representation of the collaborative workspace, where each voxel is associated with a probability of occupancy of the human, and calculates the cost based on the slowdown severity and its probability of occurrence.
Finally, the paper proposes an efficient approximation suitable for choosing among multiple equivalent goal configurations. 
This approximation reduces the computational burden in complex scenarios (Section \ref{subsec:multi-goal}).

The effectiveness of the approach is demonstrated via simulations on a three-degree-of-freedom arm (Section \ref{sec:examples}) and real experiments in a manufacturing case study (Section \ref{sec:case-study}).
A video of the experiments is also attached to the paper.

\section{Human-aware motion planning and execution}
\label{sec:background}

\begin{figure}[tpb!]
    \centering
    \includegraphics[width=.9\columnwidth]{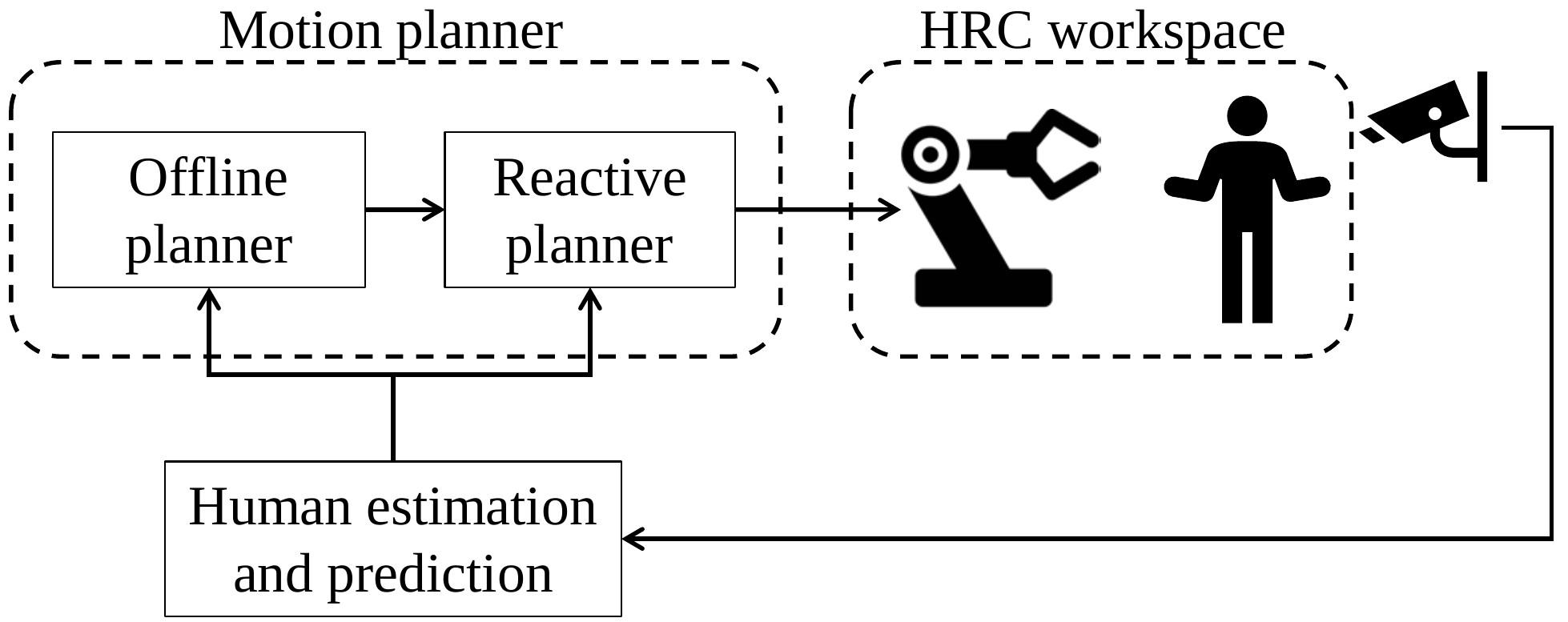}
    \caption{Human-robot collaboration planning and execution.}
    \label{fig:hrc-framework}
    \vspace{-0.45cm}
\end{figure}

Planning and execution of robot movements in a collaborative application usually follow a standard paradigm as in Figure \ref{fig:hrc-framework}.
A perception system monitors the workspace and feeds the \emph{human estimation and prediction} module that, in turn, estimates the human state at the current and near-future time.
The planning module uses this information to find trajectories that do not collide with the human and, possibly, optimize a human-centered objective.

\subsection{Motion planning}
\label{subsec:motion-planning}

Most frameworks rely on a hierarchical approach composed of an offline planner that computes collision-free trajectories and a reactive planner that modifies the motion at run-time.
The offline planner usually consists of a path planner and a path parametrization algorithm (\emph{e.g.}, TOPP \cite{pham:topp}).
Sampling-based path planners are the most widespread, for they can deal with high-dimensional search space efficiently \cite{Simeon-T-RRT, elbanhawi:sampling-based-review}.
Human awareness can be embedded in the path planner using a cost function that depends on the human state.
If trajectories are planned just before their execution, planning latency is an issue, and how to reduce it is an active research field (\emph{e.g.}, informed sampling \cite{Gammel:InformedRRT, Gammel:BIT}, anytime path planning \cite{Xu:I-FMT}).

The reactive planner modifies the trajectory during its execution to meet safety requirements and avoid unnecessary safety stops.
It can act by limiting the robot's speed, modifying the original path, or both.
In general, reactive motion planners are shifting from conservative -- yet straightforward -- strategies such as safety zones to optimized methods that adapt the robot motion continuously \cite{RCIM:dynamic-ssm, Faroni_ETFA2019}.

Notice that human-aware motion planners rely on the knowledge of the human state and, possibly, on its prediction.
Different strategies have been proposed in the literature, spanning graphical models such as Hidden Markov Models \cite{Kulic:IJRR}, recurrent neural networks \cite{fragkiadaki:RNN, Mainprice:prediction}, inverse optimal control \cite{Berret:IOC}, conditional random fields \cite{koppula2015anticipating}.
Notice that the estimation of the future human state comes with significant uncertainty, which should be explicitly taken into account in the motion planning and control modules \cite{Bajcsy:IJRR}.

\subsection{Safety requirements}
\label{subsec:safety}

Collaborative robots are required to avoid collisions with humans or limit potential damages in case of contact.
These requirements translate into limitations of the robot speed when the human is close to it.
The technical specification ISO/TS 15066 (\emph{Robots and robotic devices -- Collaborative robots}) \cite{ISOTS15066} defines speed reduction rules for collaborative operations.
If \emph{speed and separation monitoring} rules apply, then contact is not allowed.
Thus, the human-robot distance $S$ must not fall below a protective distance $S_p$.
To avoid collisions, $S_p$ is greater than or equal to the robot braking distance in the direction of the human and can be computed as:
\begin{equation}
\label{eq:ssm}
    S_p=v_h\, \bigg(T_r + \frac{v^{rh}}{a_s} \bigg) + v^{rh} T_r + \frac{{v^{rh}}^2}{2 a_s} + C,
\end{equation}

\noindent where $v^{rh}$ is the robot velocity toward the human, $v_h$ is the human velocity toward the robot, $a_s$ is the maximum Cartesian deceleration of the robot toward the human, $T_r$ is the reaction time of the robot, and $C$ is a parameter accounting for the uncertainty of the perception system.
Inequality $S\geq S_p$ translates into the limitation of the human-robot relative speed:
\begin{equation}
\label{eq:vr_leq_vmax}
    v^{rh} \leq v_{\mathrm{max}}
\end{equation}
where 
\begin{equation}
\label{eq:vmax-ssm}
    v_{\mathrm{max}}=\sqrt{v_h^2 + (a_s T_r)^2 - 2 a_s \big(C-S\big)} - a_s T_r - v_h
\end{equation}

If \emph{power and force limitation} rules apply, then contact is allowed, and the force/pressure exchanged between human and robot must be limited.
This results in the limitation $v^{rh} \leq v_{\mathrm{max}}$, where $v_{\mathrm{max}}$ depends on the robot mass and shape, and on tabular values of the maximum exchangeable force/pressure (see Annex A of \cite{ISOTS15066}).

\section{A costmap approach to human-aware planning}
\label{sec:method}

\subsection{Motivation}
\label{subsec:motivation}

The proximity of the human requires the robot to reduce its speed below a threshold.
The value of this threshold varies in time, based on the human and the robot's position and speed.
Consequently, the time taken by the robot to cover a given path not only depends on the path itself but also on the relative position and velocity of the human and the robot.
For example, given an initial and a final configuration, a short path might take more time than a long one if the short path drives the robot closer to the human.
Our goal is to embed this reasoning in the robot's path planner to search for the path that minimizes the expected execution time according to the human state and the  safety rules that will slow the robot down during the execution.

\subsection{Formulation of time-optimal human-aware path planning}
\label{subsec:optimal-path-planning}

The path planning problem consists of finding a feasible path from a starting configuration $q_{\mathrm{start}} \in \mathcal{C}$ to a set of goal configurations $Q_{\mathrm{goal}} \subset \mathcal{C}$, where $\mathcal{C}$ is the configuration space.
In robot manipulators, $q \in \mathcal{C}$ is a vector of joint positions.
The optimal path planning problem is defined as follows:
\begin{problem}
\label{problem:optimal-path-planning}
Given a starting configuration $q_{\mathrm{start}} \in \mathcal{C}$ and a set of goal configurations $Q_{\mathrm{goal}} \subset \mathcal{C}$, find a path $\sigma^*: [0, 1]\rightarrow \mathcal{C}_{\mathrm{free}}$ such that:
\begin{equation}
    \sigma^*= \argmin_{\sigma} c(\sigma) \, \, \text{s.t.} \, \, \sigma^*(0) = q_{\mathrm{start}}, \sigma^*(1) \in Q_{\mathrm{goal}}
\end{equation}
where:
    $\mathcal{C}$ is the configuration space (for robot manipulators, $q \in \mathcal{C}$ is a vector of joint positions);
    $\mathcal{C}_{\mathrm{free}} \subseteq \mathcal{C}$ is the subset given by all configurations not in collision with obstacles;
     $c$ is a positive cost function that associates a path $\sigma$ with a cost.
\end{problem}
We aim to define a cost function $c$ such that
\begin{equation}
\label{eq:execution-time}
    c = t_{\mathrm{ex}}(\sigma, \mathcal{H})
\end{equation}
where $t_{\mathrm{ex}}$ is the estimated execution time of the path, also considering the effect of the human state $\mathcal{H}$.
To do so, let $t_{\mathrm{nom}}$ be the expected execution time of a path $\sigma$ in the case that the robot does not slow down because of the human and define
\begin{equation}
    \bar{\lambda} = \frac{t_{\mathrm{ex}}(\sigma, \mathcal{H})}{t_{\mathrm{nom}}(\sigma)} \geq 1
\end{equation}
which can be seen as an average time-dilation factor that measures the effect of the human on the path  execution time.
By discretizing $\sigma$ in $w$ nodes, we can rewrite \eqref{eq:execution-time} as:
\begin{equation}
    c = \sum_{l=1}^{w-1} t_{\mathrm{nom}, l}\, \lambda\Big(\frac{q_{l+1}+q_l}{2}, \mathcal{H}\Big),
\end{equation}

\noindent where $t_{\mathrm{nom}, l}$ is the expected execution time of the segment $\overline{q_l\, q_{l+1}}$ in the case that the robot does not slow down and $\lambda$ is the time dilation factor of a segment.
$t_{\mathrm{nom}, l}$ is usually obtained a posteriori, from the path parametrization method (\emph{e.g.}, TOPP methods \cite{pham:topp}).
However, we can approximate it by imposing that at least one joint is always moving at full speed\footnote{This approximation allows decoupling the cost function from the robot's velocity. This is useful when the cost function is used within a path planner that does not consider velocities and accelerations.}.
Being $\dot{q}_{\mathrm{max}, k} > 0$ the maximum velocity of the $k$th joint, the minimum traveling time of the $k$th joint is equal to the maximum of the component-wise ratio $|q_{l} - q_{l+1}|/\dot{q}_{\mathrm{max}}$.
The minimum traveling time of the segment is therefore equal to $\| (q_{l} - q_{l+1})/\dot{q}_{\mathrm{max}} \|_{\infty}$; thus:
\begin{equation}
\label{eq:cost-function}
    c = \sum_{l=1}^{w-1} \bigg\| \frac{q_{l} - q_{l+1}}{\dot{q}_{\mathrm{max}}} \bigg\|_{\infty}  \lambda \Big(\frac{q_{l+1}+q_l}{2}, \mathcal{H}\Big)
\end{equation}

Our problem now is to derive $\lambda(\sigma, q)$ that estimates the delay factor of a configuration $q$, given the human state $\mathcal{H}$.

\begin{remark}
Cost function \eqref{eq:cost-function} may lead to an ill-posed optimization problem where solutions with equal cost but different traveling distances are equivalent.
To penalize solutions with larger travelling distance, we apply Tikhonov regularization \cite{neubauer1989tikhonov} to \eqref{eq:cost-function} and obtain:
\begin{equation}
\label{eq:cost-function-reg}
    c = \sum_{l=1}^{w-1} \bigg\| \frac{q_{l} - q_{l+1}}{\dot{q}_{\mathrm{max}}} \bigg\|_{\infty}  \lambda \Big(\frac{q_{l+1}+q_l}{2}, \mathcal{H}\Big) + \nu \sum_{l=1}^{w-1} \big\| q_{l} - q_{l+1} \big\|_{2}
\end{equation}
where $\nu>0$ is a sufficiently small regularization weight.
\end{remark}

\subsection{Cost function for deterministic representation of the human}
\label{subsec:deterministic-cost-function}

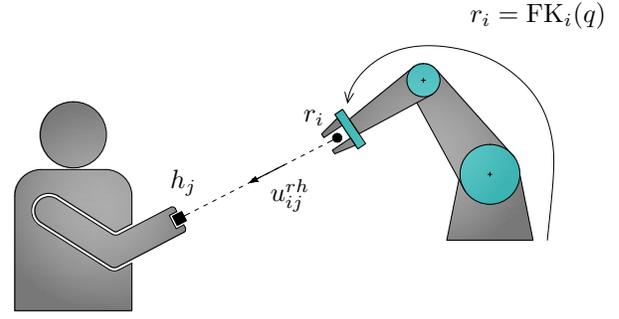
\begin{figure}[tpb]
  \centering
  \setlength{\unitlength}{0.1\columnwidth}
  \begin{picture}(10,4.8)
    \put(0,0){\includegraphics[width=0.85\columnwidth]{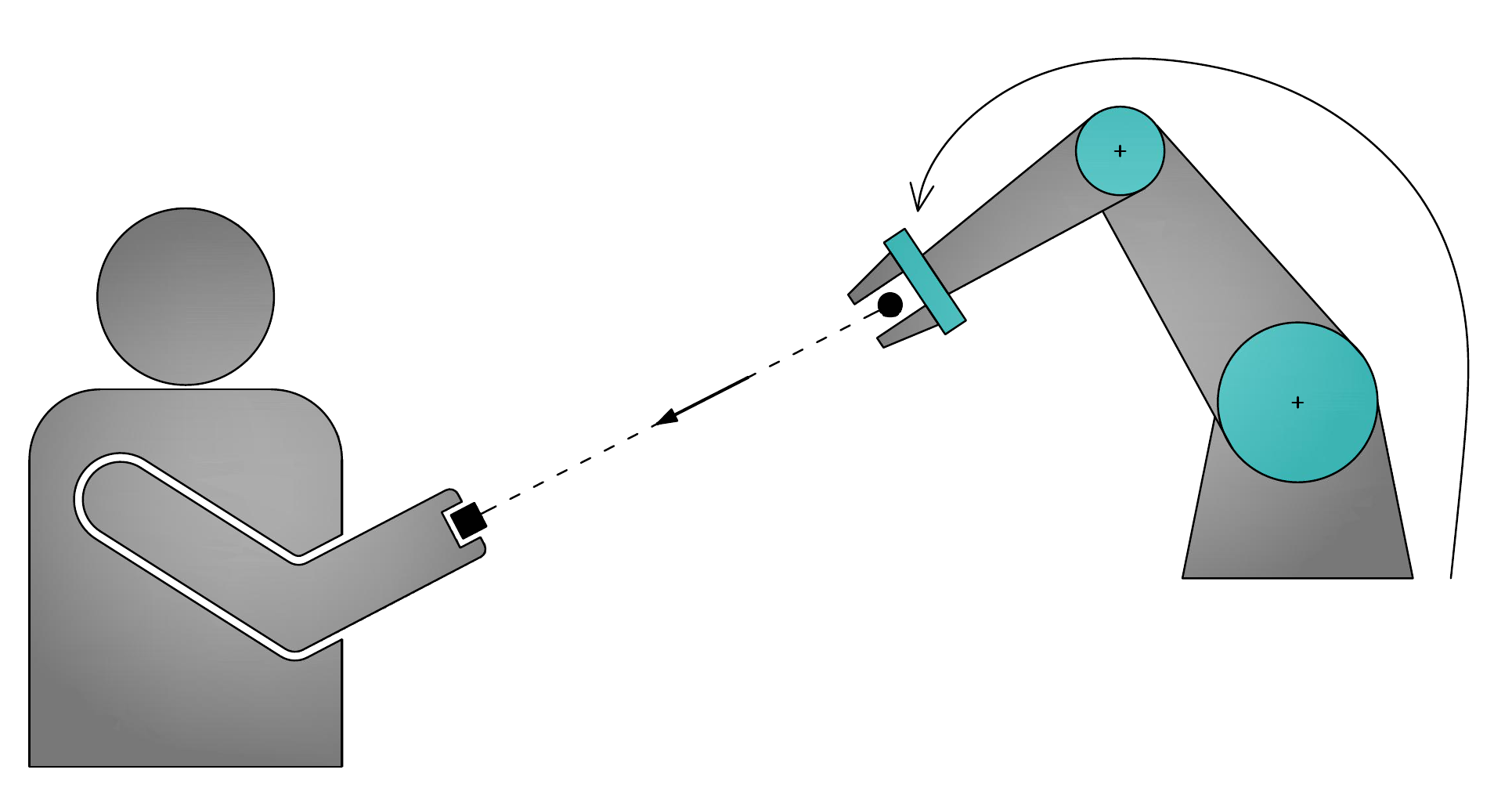}}
    \put(4.5,3){$r_i$}
    \put(2.5,2){$h_j$}
    \put(4,1.8){$u^{rh}_{ij}$}
    \put(7,4.5){$r_i=\mathrm{FK}_i(q)$}
  \end{picture}
  \caption{Graphical illustration of the variables involved in the definition of the human-aware costmap.}
  \label{fig:hrc-variables}
  \vspace{-0.4cm}
\end{figure}

Let us define the human state as a set of $m$ points, $\mathcal{H} = \{h_1, \dots, h_m\}$, where $h_j \in \mathbb R^3$ is a Cartesian point of interest of the human.
Then, consider a set of $n$ robot points, $\mathcal{R} = \{r_1, \dots, r_n\}$, $r_i \in \mathbb R^3$.
Each robot point of interest $r_i$ is related to the robot configuration $q$ through its forward kinematic function $\mathrm{FK}_i$ so that $r_i=\mathrm{FK}_i(q)$.
Let $u^{rh}_{ij}$ be the unit vector from $r_i$ to $h_j$ for all $i=1, ..., n$ and $j=1, ..., m$.
Please see Figure \ref{fig:hrc-variables} for a graphical illustration of these variables.

The maximum relative velocity between the robot and a human point $h_j$ can be expressed as:
\begin{equation}
\label{eq:relative-speed}
    v_{j}^{rh} = \max_{i=1...n}\, \, (u^{rh}_{ij})^T (\dot{r}_i-\dot{h}_j) = \max_{i=1...n} \, (u^{rh}_{ij})^T \big( J_i(q)\dot{q} - \dot{h}_j \big)
\end{equation}
where $\dot{r}$, $\dot{h}$, $\dot{q}$ are the time derivatives of $r$, $h$, and $q$, and $J_i = \frac{d\mathrm{FK_i}}{dq}$ is the linear Jacobian of $r_i$.

To express \eqref{eq:relative-speed} in $q$, we repeat the approximation made in \eqref{eq:cost-function}; that is, we impose that one joint is always moving at its maximum speed.
Being $u_l = \frac{q_{l+1}-q_{l}}{\|q_{l+1}-q_{l}\|_2}$ the unit vector of the $l$th segment of a path, it follows that:
\begin{equation}
    \dot{q}_l = K u_l\quad \text{where}\quad K = \min_k  \bigg|\frac{\dot{q}_{\mathrm{max}, k}}{u_{l, k}}\bigg| 
\end{equation}
from which:
\begin{equation}
    \label{eq:speed_scaling}
    v_{j}^{rh} =  \max_{i=1...n} \, \big(u^{rh}_{ij} \big)^T \big( J_i(q) K u_l - \dot{h}_j \big)\quad \forall \, l=1, \dots, w.
\end{equation}
$v_{j}^{rh}$ is the expected velocity at a given configuration $q$ on the $\overline{q_l\, q_{l+1}}$ segment of the path, disregarding possible safety slowdowns.
According to Section \ref{subsec:safety}, the actual robot velocity must be smaller than a maximum $v_{\mathrm{max}}$, derived from safety rules (in Speed and Separation Monitoring, $v_{\mathrm{max}}$ is given by \eqref{eq:vmax-ssm}).
If $v_{j}^{rh} \leq v_{\mathrm{max}}$ $\forall j$, speed and traveling time are invariate.
Otherwise, a speed reduction ratio equal to $\min_j (v_{\mathrm{max}} / v_{j}^{rh})$ is expected, leading to an increase of the traveling time of a factor equal to $\max_j (v_{j}^{rh} / v_{\mathrm{max}}$). 
Therefore, the time dilation function $\lambda$ of a given configuration $q$ results in:
\begin{equation}
\label{eq:lamda-deterministic}
    \lambda(q, \mathcal{H}) = \max\big(\, \lambda_1, \dots, \lambda_m, 1\, \big) 
\end{equation}
where $\lambda_j = \max_j \frac{v_{j}^{rh}}{v_{\mathrm{max}}}$.
The time-optimal human-aware cost function $c$ is finally obtained using \eqref{eq:lamda-deterministic} in \eqref{eq:cost-function}.\\
\emph{Note:} The velocity of the human points, $\dot{h}_j$, can hardly be foreseen with low uncertainty during the planning phase. In the experiments, we set $\dot{h}_j=0$ in \eqref{eq:speed_scaling} and $v_h=0$ in \eqref{eq:vmax-ssm}.

\begin{figure}[tpb]
    \includegraphics[width=0.45\columnwidth]{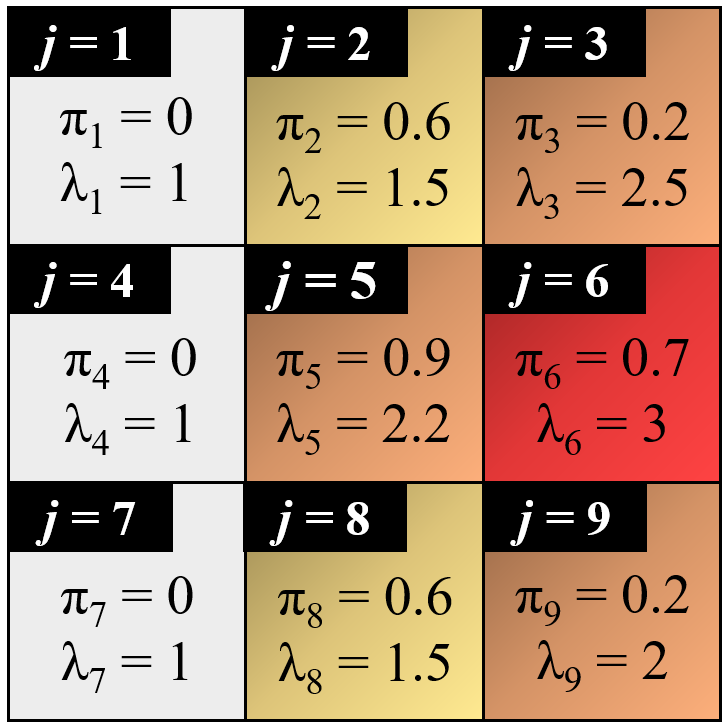} 
    \llap{\makebox[0pt][c]{\raisebox{2cm}{\hspace{3.8cm} $\hat{\lambda} = \begin{bmatrix} \lambda_6\\ \lambda_3 \\ \lambda_5 \\ \lambda_9 \\ \lambda_2 \\ \lambda_8 \\ \lambda_1 \\ \lambda_4 \\ \lambda_7 \end{bmatrix} \quad\hat{\pi} = \begin{bmatrix} \pi_6\\ \pi_3 \\ \pi_5 \\ \pi_9 \\ \pi_2 \\ \pi_8 \\ \pi_1 \\ \pi_4 \\ \pi_7 \end{bmatrix}$ }}}
    \caption{Illustrative example of 9 voxels with their values of $\pi_j$ and $\lambda_j$. $\lambda$ results from \eqref{eq:lambda-probabilistic}, and it is sorted in descending order, while $\hat{\pi}$ have elements $\pi_j$ sorted according to $\hat{\lambda}$.}
    \label{fig:example-lambda}
\end{figure}

\subsection{Cost function for probabilistic representation of the human}
\label{subsec:probabilistic-cost-function}

We aim to derive the time dilation factor $\lambda(q, \mathcal{H})$ if the human points of interest are expressed as a probability distribution function.
Let us discretize the Cartesian workspace in a set of voxel and define the voxelization function
\begin{equation}
    v\, :\, \mathbb R^3 \rightarrow V=\{1, \dots, m\}\subset \mathbb N, \, \, x\mapsto j
\end{equation}
which associates the point $x$ with the voxel $j$.
We define the human state as a sequence of probability distributions, $\mathcal{H}=\{\pi_1, ..., \pi_m\}$, where $\pi_j \in [0, 1]$ is the expected value of the probability of voxel $j$ to be occupied by a human.

Denoted by $C_j$ the center of voxel $j$, we can compute $\lambda(q, C_j)$ such that $\pi_j>0$.
If $\pi_j=0$ then $\lambda_j=1$.
For each configuration $q$ and voxel $j$, we compute an expected occupancy probability $\pi_j$ and a time-dilation factor $\lambda_j$.

To derive the expected value of $\lambda$, we need to bear in mind that, in case two voxels are occupied at the same time, the realized value of $\lambda$ is the maximum of the two (\emph{i.e.} the worst case). When the worst-case voxel is occupied, it determines the value of $\lambda$ regardless of the occupancy of the others.
If the worst-case voxel is not occupied, the same reasoning applies to the second worst-case voxel, and so on.

To have a clear understanding of this, consider the example of Figure \ref{fig:example-lambda}.
The example shows $m=9$ voxels, and voxel $j=6$ has the higher value of $\lambda_j$.
The probability that $\lambda = \lambda_6$ is equal to the probability of voxel $j$ to be occupied, regardless of the occupation of the other voxels, \emph{i.e.}
$
    P(\lambda = \lambda_6) = \pi_6.
$

The probability of the second most critical $\lambda_j$ to occur, $P(\lambda=\lambda_3)$, is equal to the probability of voxel $3$ to be occupied, conditioned to the probability of voxel $6$ not being occupied
$
    P(\lambda = \lambda_3) = \pi_3 (1-\pi_6).
$

To generalize this reasoning, let $\hat{\lambda}$ be a vector of $\{\lambda_j\}$ sorted in descending order, and $\hat{\pi}$ be a vector of $\{\pi_j\}$ ordered according to $\hat{\lambda}$.
The probability that $\hat{\lambda}_y$ occurs is:
\begin{equation}
    P(\lambda=\hat{\lambda}_y) =
    \begin{cases}
    \hat{\pi}_1 & \text{if $y=1$} \\
    \hat{\pi}_y\, \prod_{k=1}^{y-1} \big(1-\hat{\pi}_k\big) & \text{if $y \in \{2, \dots, m\}$}
  \end{cases}
\end{equation}
Moreover, the probability that no voxel will be occupied is equal to
$\prod_{k=1}^m \big(1-\hat{\pi}_k\big)$.
In that case, $\lambda$ would be equal to 1.

The expected value of $\lambda$ is therefore the average value of $\hat{\lambda}_y$ weighted by its probability to occur:
\begin{equation}
\label{eq:lambda-probabilistic}
    \lambda = \sum_{y=1}^m \hat{\lambda}_y\, P(\lambda=\hat{\lambda}_y) + \prod_{y=1}^m \big( 1-\hat{\pi}_y \big)
\end{equation}
Finally, time-optimal human-aware cost function $c$ results by using \eqref{eq:lambda-probabilistic} in \eqref{eq:cost-function}.

\subsection{An approximation for multiple goals}
\label{subsec:multi-goal}

A path planning problem can involve multiple goals.
A practical example is when multiple configurations correspond to the same robot pose or when multiple instances of the same objects are available in the scene.
A large number of goals increases the number of evaluations of cost function \eqref{eq:cost-function}\footnote{In the worst case, the computational time is linear in the number of goals. This occurs when the planning problem is solved for each goal independently.}.
Considering that a sampling-based path planner can take up to some seconds in real-world problems, the planning time may soon become unbearable for online purposes.
When planning latency is an issue, it is helpful to approximate \eqref{eq:cost-function} to reduce the number of evaluations of $\lambda(q, \mathcal{H})$.
The proposed approximation assumes that the cost of goal configurations is predominant in selecting human-aware paths when multiple goals are available.
For this reason, we aim to minimize a multi-objective cost function composed of the path length and a terminal human-aware cost:
\begin{equation}
\label{eq:multi-goal-cost-function}
    c(\sigma, \mathcal{H}) = \sum_{k=1}^{w-1} \| q_{k+1} - q_k \|_2 + b \, \lambda(q_w, \mathcal{H})
\end{equation}
where $b\geq0$ is a weighting factor and $q_w \in Q_{\mathrm{goal}}$.
Compared with \eqref{eq:cost-function}, \eqref{eq:multi-goal-cost-function} only requires the computation of $\lambda$ for all goal configurations.
As a drawback, the optimal solution is a minimum-length path toward the goal that minimizes \eqref{eq:multi-goal-cost-function}; \emph{i.e.}, the problem neglects the human effect on the path waypoints.

\subsection{Trajectory execution}
\label{subsec:execution}

The proposed approach allows path planners to find optimal human-aware paths.
Then, the path is parametrized through a time-optimal path parametrization method \cite{pham:topp} and executed by the robot.
The execution phase needs to consider safety specifications, as mentioned in \ref{subsec:safety}.
According to Speed and Separation Monitoring rules, the nominal robot speed is modified in real-time by computing a speed scaling factor $s_{\mathrm{ovr}} \in [0, 1]$ chosen as follows:
\begin{equation}
\label{eq:s_ovr}
    s_{\mathrm{ovr}} = \min \bigg( \frac{v_{\mathrm{max}}}{{v_{\mathrm{max}}^{rh}}} \, , \, 1 \bigg)
\end{equation}
where $v_{\mathrm{max}}$ is calculated from \eqref{eq:vmax-ssm} and, recalling \eqref{eq:relative-speed}, 
$$
v_{\mathrm{max}}^{rh} = \max_j v_j^{rh}
$$
\noindent is the maximum relative speed allowed at the current instant.

\section{Numerical examples}
\label{sec:examples}


This section considers a toy problem in which a 3-degree-of-freedom manipulator performs point-to-point movements.
Starting and goal configurations are sampled randomly from the robot configuration space.
A voxel-based representation models the human occupancy, $
\mathcal{H}=\{\pi_1, ..., \pi_m\},
$
where $\pi_j \in [0, 1]$ is the expected value of the probability of voxel $j$ to be occupied by a human and is assigned according to the following function $f_{\pi}$:
\begin{equation}
    f_{\pi} = 
    \begin{cases}
    1-r\, \| \mu - c_j \|_2 & \text{if $r\, \| \mu - c_j \|_2 \leq 1$} \\
    0 & \text{otherwise}
  \end{cases}
\end{equation}
where $\mu \in \mathbb R^3$ is a random Cartesian point in the robot workspace, $c_j \in \mathbb R^3$ is the center of voxel $j$, and $r=0.5$~m is a user-selected occupancy radius.


\begin{table*}[tpb]
	\caption{Numerical results, normalized by MIN-PATH values (100 random trials repeated 6 times for each planner).}	
	\label{tab:hamp_results}
	\centering
\makebox[0.9\textwidth]
{
	\centering
	\subfloat[][Experiment A (static human position), $C=0.2$ m]
	{
		$
		\begin{array}{lc}
		\toprule
		& \text{HAMP -- mean (std.dev.)} \\
		\midrule
		\text{Path length}          & \textbf{1.30}     (0.141)\\
		\text{Execution time}       & \textbf{0.81}     (0.092)\\
		\text{Safety speed delay}   & \textbf{0.83}     (0.092) \\
		\midrule
		\text{Success rate } & \text{HAMP = } \textbf{0.985} \quad\text{MIN-PATH = } 0.901 \\
		\bottomrule
		\end{array}
		$
	\label{tab:hamp_results_02m}
	}\hfill
	\subfloat[][Experiment A (static human position), $C=0.5$ m]
	{
		$
		\begin{array}{lc}
		\toprule
		& \text{HAMP -- mean (std.dev.)} \\
		\midrule
		\text{Path length} &  \textbf{1.37}(0.187) \\
		\text{Execution time} & \textbf{0.89}(0.121)  \\
		\text{Safety speed delay} & \textbf{0.82}(0.112) \\
		\midrule
		\text{Success rate } & \text{HAMP = } \textbf{0.829} \quad \text{MIN-PATH = } 0.675 \\
		\bottomrule
		\end{array}
		$
		\label{tab:hamp_results_05m}
		}
}
\\
\makebox[0.9\textwidth]
{
	\centering
	\subfloat[][Experiment B (probabilistic update of the human position)]
	{
		$
		\begin{array}{lcc}
		\toprule
		& \text{HAMP} & \text{HAMP-Probabilistic} \\
		& \text{mean (std.dev.)} & \text{mean (std.dev.)} \\
		\midrule
		\text{Path length} & 1.32(0.256) & \textbf{1.39}(0.224) \\
		\text{Execution time} & 0.90(0.220) & \textbf{0.89}(0.161) \\
		\text{Safety speed delay} & 0.73(0.200) & \textbf{0.69}(0.154) \\
		\text{Success rate } & 0.86 & \textbf{0.93} \\
		\bottomrule
		\end{array}
		$
		\label{tab:hamp_results_probabilistic}
	}\hfill
	\subfloat[][Experiment C (multiple equivalent goals)]
	{
		$
		\begin{array}{lcc}
		\toprule
		& \text{HAMP} & \text{HAMP-Approximated} \\
		& \text{mean (std.dev.)} & \text{mean (std.dev.)} \\
		\midrule
		\text{Path length} & 1.87(0.440) & \textbf{1.77}(0.458) \\
		\text{Execution time} & 0.49(0.147) & \textbf{0.52}(0.181) \\
		\text{Safety speed delay} & 0.24(0.072) & \textbf{0.29}(0.102) \\
		\text{Success rate } & 0.99 & \textbf{0.98} \\
		\bottomrule
		\end{array}
		$
		\label{tab:hamp_results_multigoal}
	}
}
\vspace{-0.5cm}
\end{table*}

\begin{figure*}[t!]
	\centering
	\subfloat[][Experiment A, $C=0.2$ m]
	{\includegraphics[trim = 0cm 0.2cm 0cm 1.45cm, clip, angle=0, width=\columnwidth]{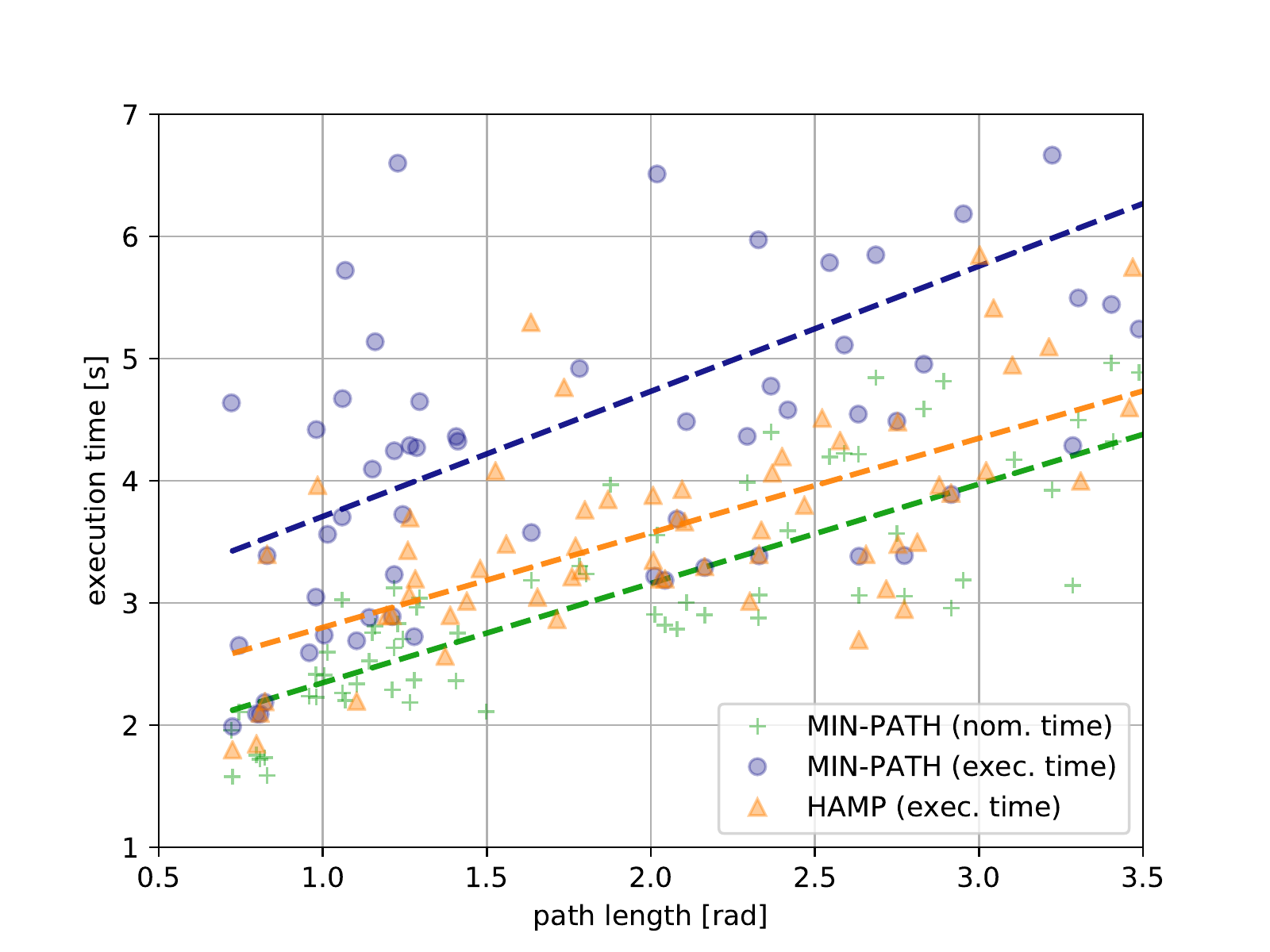}
	\label{fig:path_vs_time_02m}}\hfill
	\subfloat[][Experiment A, $C=0.5$ m]
	{\includegraphics[trim = 0cm 0.2cm 0cm 1.45cm, clip, angle=0, width=\columnwidth]{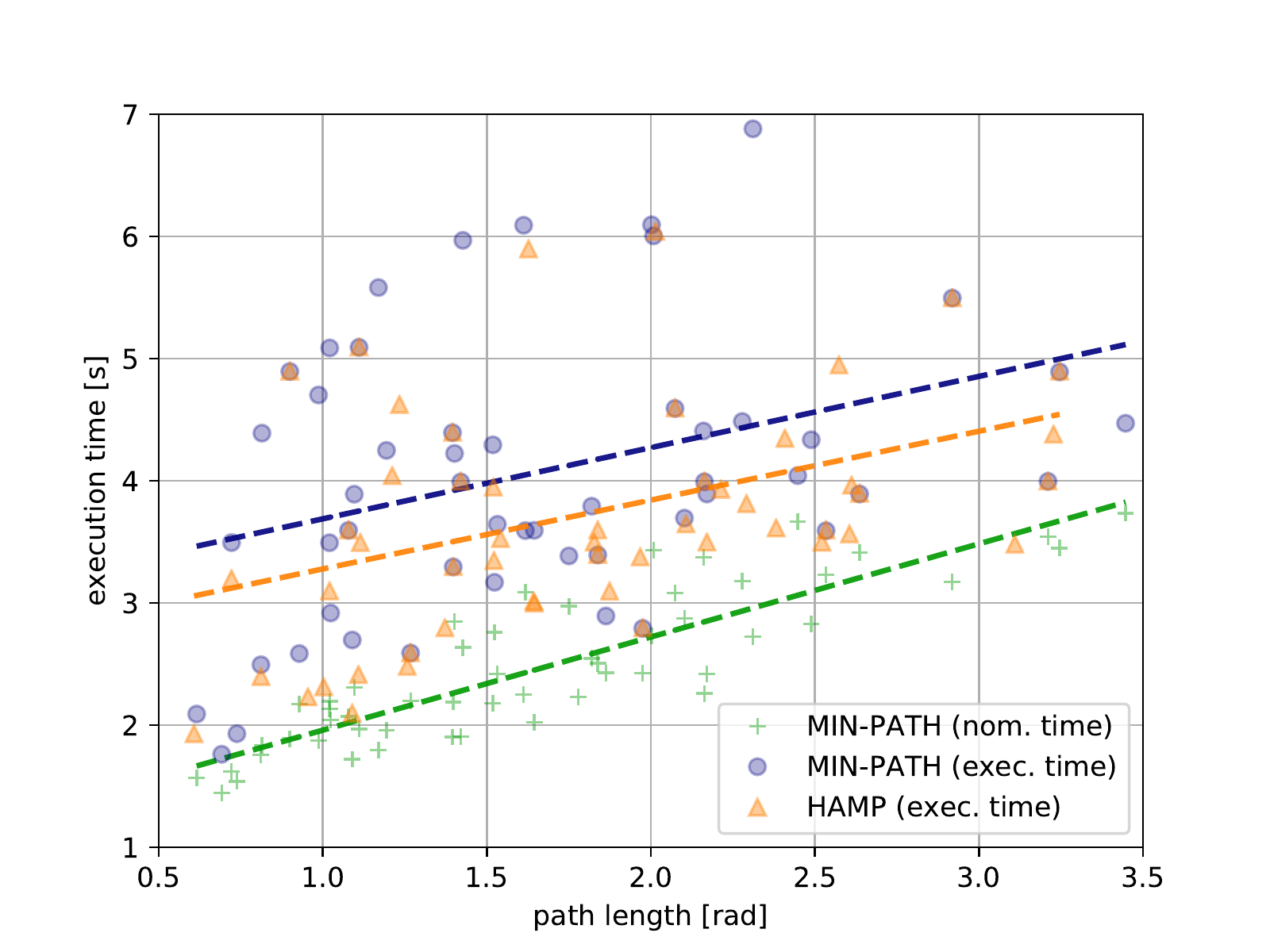}
	\label{fig:path_vs_time_05m}}
	\caption{Lenght and execution time of each trajectory of Experiment A. Green crosses: MIN-PATH with nominal execution time (\emph{i.e.}, disregarding safety slowdowns); Blue circles: MIN-PATH with actual execution time; Orange triangles: HAMP with actual execution time. Dashed lines are the linear regression of the respective colors.}
	\label{fig:path_vs_time}
	\vspace{-0.3cm}
\end{figure*}

\subsection{Experiment A: human-aware motion planning assessment}

In the first experiment, we assess the effectiveness of the human-aware motion planner (HAMP) compared to non-human-aware approaches in case the human position is static (\emph{i.e.}, the human position is kept constant during each query).
We compare the following motion planners: a) MIN-PATH, which solves a minimum-length path planning problem disregarding the human presence; b) HAMP: minimization of the expected execution time with human awareness \eqref{eq:cost-function-reg}.
Both planners use Informed-RRT$^*$ \cite{Gammel:InformedRRT} to solve the planning problem.
Both implement the trajectory execution module (Section \ref{subsec:execution}) to prevent collisions if the path collides with the worker. 
The trajectory execution module is a possible implementation of a safety module that slows the robot down based on the human-robot relative state.
It is utterly necessary during real experiments to avoid the robot collides with the worker.

We generate 100 planning queries; each query is solved and executed 6 times.
We measure the path length, the execution time, the time delay owed to safety slowdowns (calculated as the ratio between the actual and the planned execution time), and whether the execution succeeded or failed (failures include the inability to find a path and safety stops).
To compare queries, we normalize the results of each query obtained by HAMP to the median value of the results obtained by MIN-PATH, considered as the baseline.
We run two sets of trials with different safety distance thresholds $C$\footnote{According to \eqref{eq:vmax-ssm}, $C$ is a safety threshold that accounts for uncertainty in perception and actuation, \textit{i.e.}, the safety margin from the human closest point.}.

Results for $C=0.2$ m are in Table \ref{tab:hamp_results_02m}.
HAMP finds longer paths (+30\% on average) than MIN-PATH. 
Nonetheless, the average execution time and the average safety delay are significantly smaller (-19\% and -17\%) because HAMP's paths are less prone to interfere with the operator. 
Remarkably, HAMP improves the success rate by 9\%, thanks to its ability to avoid areas too close to the operator.
Results for $C=0.5$ m are in Table \ref{tab:hamp_results_05m}.
This second set of trials is interesting to understand how the human-awareness behaves with more severe safety implementations.
As for the previous trials, HAMP finds longer paths (+37\% on average) with shorter average execution times (-11\%).
As the safety threshold is larger, more trajectory interruptions are expected to happen.
HAMP adapts to this change by increasing the path lengths, leading to an improvement in the success rate of around +23\%.
Interestingly, a more conservative implementation of safety functions mainly impacts the success rate of HAMP.

To better clarify the behavior of HAMP, Figure \ref{fig:path_vs_time} plots all queries by their path length and execution time.
As expected, the nominal execution time for MIN-PATH's trajectories (green line) lies in the bottom part of the graph as it neglects the effect of safety slowdowns.
The same trajectories, plotted versus their actual execution time (blue), show a large offset and a larger dispersion due to the activation of the safety slowdown.
On the contrary, HAMP's trajectories lie significantly below MIN-PATH's, \textit{i.e.}, the execution time is shorter than that of MIN-PATH, being equal to the path length.
Notice that the slowdowns are caused by the trajectory execution module (Section \ref{subsec:execution}), which was used with both planners to prevent collisions.
Therefore, HAMP implies less severe safety speed reduction during execution.

\begin{figure}[tpb]
	\centering
	\includegraphics[trim = 0cm 0cm 0cm 1.45cm, clip, angle=0, width=0.95\columnwidth]{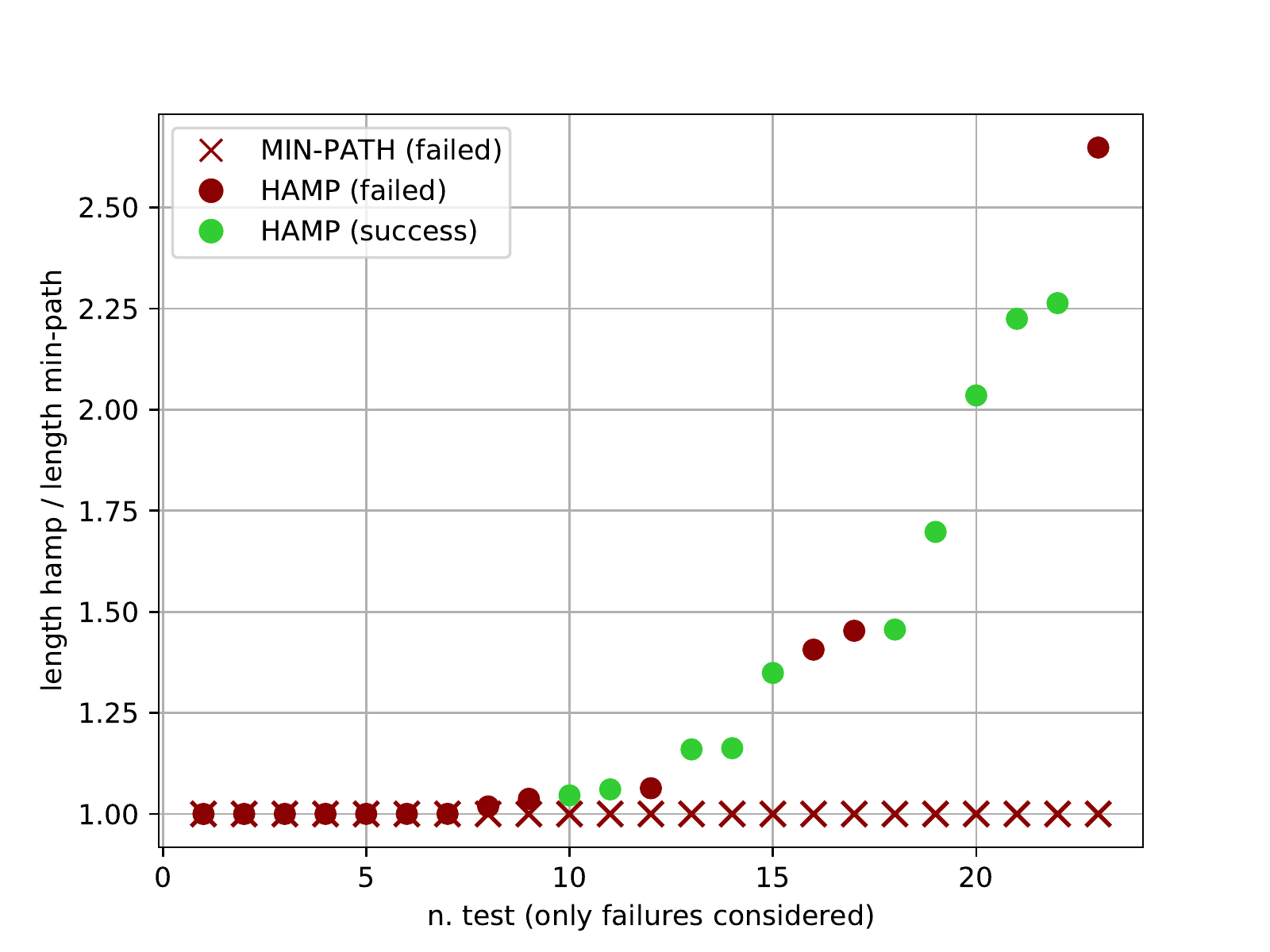}
	\caption{Failures during execution by at least one planner (Exp. A, $C=0.5$ m). 
	Crosses: MIN-PATH; Circles: HAMP. Red markers indicate a failure, green markers indicate a success.}
    \label{fig:failures-visual}		
    \vspace{-0.3cm}
\end{figure}

Speaking of failures, it is also interesting that HAMP can perform the trajectory when MIN-PATH fails.
Figure \ref{fig:failures-visual} shows the path lengths of HAMP's trajectories for all the queries where MIN-PATH failed.
Red and green markers indicate a failure or a success; notice that green circles (\emph{i.e.}, HAMP successes) are more likely when HAMP's lengths are larger.
Remarkably, during our experiments, MIN-PATH could never execute a query at which HAMP failed.

\subsection{Experiment B: probabilistic human-aware motion planning assessment}

In this second experiment, we update the human position during the experiment according to the probability distributions $\mathcal{H}$. 
This is useful to highlight the advantages of the probabilistic formulation of the human-aware motion planner given in Section~\ref{subsec:probabilistic-cost-function}.
Indeed, existing human-aware motion planners assume deterministic representation of the human's state and prediction \cite{Alami:hamp, Mainprice:hamp}.

We compare the following motion planners: 
i) MIN-PATH: minimization of the path length; ii) HAMP: minimization of the expected execution time with deterministic human awareness (\emph{i.e.}, \eqref{eq:cost-function-reg} with $\lambda$ computed as described in Section~\ref{subsec:deterministic-cost-function}); iii) HAMP-Probabilistic: minimization of the expected execution time with probabilistic human awareness (\emph{i.e.}, \eqref{eq:cost-function-reg} with $\lambda$ computed as described in Section~\ref{subsec:probabilistic-cost-function}).
We generate 100 queries, and each query is executed 6 times by each planner.

Results are in Table \ref{tab:hamp_results_probabilistic}.
Both human-aware methods lead to longer paths, shorter execution times, and higher success rates.
In general, HAMP-Probabilistic gives better results, especially considering the success rate.
Compared to the first experiment, HAMP's success rate worsens because it is not able to predict the future human position.
As a consequence, the new position of the human is more often on the way of the robot's path.
On the contrary, HAMP-Probabilistic's success rate is comparable to that of the first experiment because the planner accounts for the future possible position of the human.

\subsection{Experiment C: Assessment of the multi-goal approximation}

This experiment assesses the validity of the approximated multi-goal approach described in Section~\ref{subsec:multi-goal}.
We set up an experiment where the goal of each query is a set of 20 randomly sampled configurations.
We compare the following motion planners: i) MIN-PATH: minimization of the path length; ii) HAMP: minimization of \eqref{eq:cost-function-reg}; iii) HAMP-Approximated: minimization of  \eqref{eq:multi-goal-cost-function}.
As in the previous experiments, we generate 100 random queries, and each query is executed six times by each planner.
Results are in Table \ref{tab:hamp_results_multigoal}.
Both human-aware approaches dramatically reduce the execution time (HAMP: -50\%, Hamp-Approximated: -45\%) by finding much longer paths (HAMP: +100\%, Hamp-Approximated: +70\%).
This improvement is because human-aware planners often find a goal that does not require safety slowdowns when multiple equivalent goals are available.
For example, if a goal $q_{\mathrm{goal}1}$ is further than a goal $q_{\mathrm{goal}2}$ but closer to the human, both HAMP and HAMP-Approximated would probably choose the second goal, while MIN-PATH would still go for the first one.

The tests prove the validity of the approximation \eqref{eq:multi-goal-cost-function}, as it yields results similar to HAMP. 
This demonstrates that, in the multi-goal case, the human-awareness cost of the goal configuration is often higher than that of the whole path.

\section{Case study}
\label{sec:case-study}

\begin{figure}[tpb]
	\centering
	\includegraphics[trim = 0cm  0.2cm 0cm 0.6cm, clip, angle=0, width=.7\columnwidth]{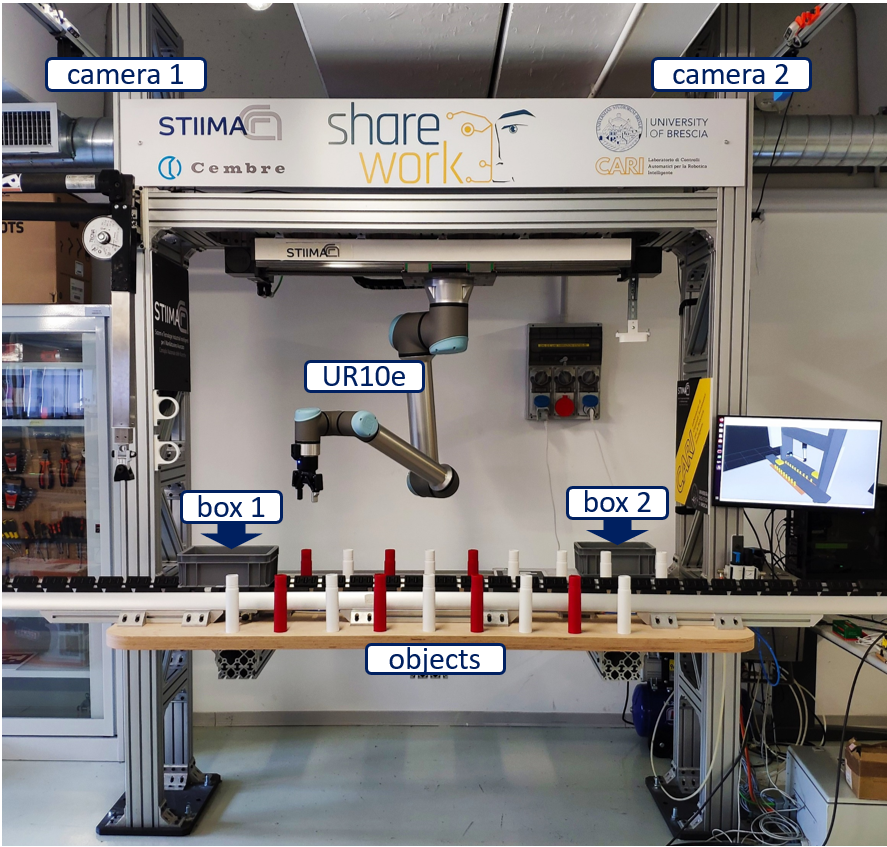}
	\caption{Experimental setup. The robot picks and places objects from and into the boxes on the panel, while the human works in the workspace.}
	\label{fig:foto-cella}	
\end{figure}

We tested our approach on a cell designed within the EU-funded project \emph{Sharework}.
As shown in Figure \ref{fig:foto-cella}, the cell is composed of a $6$-degree-of-freedom collaborative robot, Universal Robots UR10e, mounted upside down.
It is equipped with two fixed cameras, Intel Realsense D435, with a frame rate equal to 30 Hz, acting as a perception system for the human position.
The case study is a collaborative process where the robot should pick 16 objects from the table and place them in one of the two boxes.
In the meantime, the operator assembles mechanical parts near the robot.
%
%
We implement a planning and control architecture such as that of Figure \ref{fig:hrc-framework}.
The offline planner computes a path just before every robot's movement.
Once an optimal path has been found (or the maximum planning time has expired), the path is executed according to Section \ref{subsec:execution}.
We use a point-cloud representation of the human and feed the human prediction module with the centroid of the point cloud.
We use a simple prediction model that assumes a constant human position based on the moving average of the centroid position over the last second.

\subsection{Experiments and results}
\label{sec:results-experimental}

We compared MIN-PATH (minimization of the path length) and HAMP-Approximated (minimization of the expected execution time with goal-approximated human awareness \eqref{eq:multi-goal-cost-function}).
The experiments involved $10$ participants\footnote{All subjects participated voluntarily, signing an informed consent form in accordance with the Declaration of Helsinki.}.
Each participant performed the task twice: once when the robot was using MIN-PATH and once when using HAMP-Approximated.
The participants were unaware of which motion planner was running during each test; the order of the motion planners was chosen randomly.
For each test, we computed four indices: the time taken by the robot to complete all the pick\&place tasks (in seconds); the time taken by the operator  to complete all the pick\&place tasks (in seconds); the average speed scaling (\% of the nominal speed); and the average distance between the robot and the human (in meters).
Results are in Table~\ref{tab:hamp_results_experiments}.
With HAMP-Approximated, the overall execution time is shorter (-19\%) and the average speed scaling is greater (+16\%). 
The average distance between the human and the robot obtained using HAMP-Approximated is around 11\% greater than that obtained using MIN-PATH.
Results do not show a significant difference in the operators' execution time, suggesting that the choice of the planner do not significantly affect the operators' behavior.
The experiments are shown in the video attached to this paper.
\rev{Notice that the order of the pick\&place tasks is chosen by the planners.
HAMP-Approximated is able to find the sequence that reduces interference with the human, leading to less severe slowdowns.
}

\begin{table}[tpb]
    \caption{Experimental results: execution time, average speed scaling, and the average distance between the human and the robot for 16 pick\&place operations. 
	}	
	\label{tab:hamp_results_experiments}
	\centering
	$
	\begin{array}{lcc}
	\toprule
	& \text{HAMP-Approximated} & \text{MIN-PATH} \\
	& \text{mean (std.dev.)} & \text{mean (std.dev.)} \\
	\midrule
    \text{Robot execution time [s]}  & \textbf{135}(15.0)   & 166(23.9) \\
    \text{Human execution time [s]}  & \textbf{109}(17) & 115(13) \\
    \text{Safety speed scaling [\%]} & \textbf{91.2}(2.67)  & 78.8(18.6) \\
    \text{Distance human-robot [m]}  & \textbf{1.07}(0.074) & 0.96(0.15) \\
	\bottomrule
	\end{array}
	$
 \vspace{-0.3cm}
\end{table}

\subsection{Discussion}
\label{sec:experimental-discussion}

Experimental results are coherent with those obtained in simulation in Section~\ref{sec:examples}. However, the performance enhancement measured in real-world tests is less accentuated than that obtained with simulations.
The human-aware motion planner reduced the execution time by 45\% in simulation and 19\% in real tests (Table~\ref{tab:hamp_results_multigoal} and \ref{tab:hamp_results_experiments}).
Although this difference partially owes to the differences between the two scenarios (\emph{i.e.}, different robot and environment, different number of available goals), a key aspect is the different behavior of the human state.
In simulated tests, the human kept still throughout each planning and execution (Experiment A) or updated according to the probability distribution (Experiment B).
In real-world tests, we used a simplified model to predict human movements, which could include large whole-body movements across the cell (see the attached videos).
Therefore, the accuracy of the human's state prediction is utterly important, and the planning scheme of Figure~\ref{fig:hrc-framework} may benefit from embedding the proposed approach into an online path re-planner \cite{Tonola_ROMAN2021}.

\section{Conclusions}
\label{sec:conclusions}

This paper proposed a novel method to embed human awareness into robot path planners.
Compared with existing approaches, our method explicitly addresses minimization of the trajectory execution time considering the safety speed reduction owed to the proximity of the human and consider the uncertainty of the human estimation in estimating the robot slowdown.
As demonstrated in simulations and real-world experiments, these features significantly reduce execution time and avoid unnecessary safety speed reduction.
Future works will aim to reduce the  computational time to calculate the time-dilation factor $\lambda$ by directly estimating the human-robot distance or the time dilation factor through an artificial neural network.
The aim is two-fold: allow for the integration of the approach into path re-planning algorithms, and use larger values of $n$ and $m$ to model the robot and the human (regarding the second point, notice that the computation of $\lambda$ for multiple points can be easily parallelized).

\bibliographystyle{IEEEtran}
\bibliography{references, reference_stiima}             

\begin{thebibliography}{10}
\providecommand{\url}[1]{#1}
\csname url@samestyle\endcsname
\providecommand{\newblock}{\relax}
\providecommand{\bibinfo}[2]{#2}
\providecommand{\BIBentrySTDinterwordspacing}{\spaceskip=0pt\relax}
\providecommand{\BIBentryALTinterwordstretchfactor}{4}
\providecommand{\BIBentryALTinterwordspacing}{\spaceskip=\fontdimen2\font plus
\BIBentryALTinterwordstretchfactor\fontdimen3\font minus
  \fontdimen4\font\relax}
\providecommand{\BIBforeignlanguage}[2]{{%
\expandafter\ifx\csname l@#1\endcsname\relax
\typeout{** WARNING: IEEEtran.bst: No hyphenation pattern has been}%
\typeout{** loaded for the language `#1'. Using the pattern for}%
\typeout{** the default language instead.}%
\else
\language=\csname l@#1\endcsname
\fi
#2}}
\providecommand{\BIBdecl}{\relax}
\BIBdecl

\bibitem{Umbrico:ontology}
A.~Umbrico, A.~Orlandini, and A.~Cesta, ``An ontology for human-robot
  collaboration,'' \emph{Procedia CIRP}, vol.~93, pp. 1097--1102, 2020.

\bibitem{Makris:task-planning-hrc}
G.~Michalos, J.~Spiliotopoulos, S.~Makris, and G.~Chryssolouris, ``A method for
  planning human robot shared tasks,'' \emph{CIRP J. Manuf. Sci. Technol.},
  vol.~22, pp. 76--90, 2018.

\bibitem{Makris:scheduling-hrc}
N.~Nikolakis, N.~Kousi, G.~Michalos, and S.~Makris, ``Dynamic scheduling of
  shared human-robot manufacturing operations,'' \emph{Procedia CIRP}, vol.~72,
  pp. 9--14, 2018.

\bibitem{Lasota:hamp}
P.~A. Lasota and J.~A. Shah, ``Analyzing the effects of human-aware motion
  planning on close-proximity human--robot collaboration,'' \emph{Human
  factors}, vol.~57, no.~1, pp. 21--33, 2015.

\bibitem{haw-control}
J.~de~Gea~Fern{\'a}ndez, D.~Mronga, M.~G{\"u}nther, T.~Knobloch, M.~Wirkus,
  M.~Schr{\"o}er, M.~Trampler, S.~Stiene, E.~Kirchner, V.~Bargsten
  \emph{et~al.}, ``Multimodal sensor-based whole-body control for human--robot
  collaboration in industrial settings,'' \emph{Rob. Auton. Syst.}, vol.~94,
  pp. 102--119, 2017.

\bibitem{SAVERIANO201796}
M.~Saveriano, F.~Hirt, and D.~Lee, ``Human-aware motion reshaping using
  dynamical systems,'' \emph{Pattern Recognit. Lett.}, vol.~99, pp. 96--104,
  2017.

\bibitem{Tonola_ROMAN2021}
C.~Tonola, M.~Faroni, N.~Pedrocchi, and M.~Beschi, ``Anytime informed path
  re-planning and optimization for human-robot collaboration,'' in \emph{Proc.
  IEEE Int. Conf. on Robot and Human Interactive Comm.}, 2021.

\bibitem{Zanchettin:safety}
A.~M. {Zanchettin}, N.~M. {Ceriani}, P.~{Rocco}, H.~{Ding}, and B.~{Matthias},
  ``Safety in human-robot collaborative manufacturing environments: Metrics and
  control,'' \emph{IEEE Trans. Autom. Sci. Eng.}, vol.~13, no.~2, pp. 882--893,
  2016.

\bibitem{RCIM:implementing-ssm}
J.~A. Marvel and R.~Norcross, ``Implementing speed and separation monitoring in
  collaborative robot workcells,'' \emph{Rob. Comput. Integr. Manuf.}, vol.~44,
  pp. 144--155, 2017.

\bibitem{RCIM:dynamic-ssm}
C.~Byner, B.~Matthias, and H.~Ding, ``Dynamic speed and separation monitoring
  for collaborative robot applications--concepts and performance,'' \emph{Rob.
  Comput. Integr. Manuf.}, vol.~58, pp. 239--252, 2019.

\bibitem{Palleschi:safe-time-optimal}
A.~Palleschi, M.~Hamad, S.~Abdolshah, M.~Garabini, S.~Haddadin, and
  L.~Pallottino, ``Fast and safe trajectory planning: Solving the cobot
  performance/safety trade-off in human-robot shared environments,'' \emph{IEEE
  Rob. Autom. Lett.}, vol.~6, no.~3, pp. 5445--5452, 2021.

\bibitem{Zanchettin:proactive-motion-planning}
A.~{Casalino}, D.~{Bazzi}, A.~M. {Zanchettin}, and P.~{Rocco}, ``Optimal
  proactive path planning for collaborative robots in industrial contexts,'' in
  \emph{IEEE Int. Conf. on Robotics and Automation}, 2019, pp. 6540--6546.

\bibitem{Alami:hamp}
E.~A. {Sisbot} and R.~{Alami}, ``A human-aware manipulation planner,''
  \emph{IEEE Transactions on Robotics}, vol.~28, no.~5, pp. 1045--1057, 2012.

\bibitem{Mainprice:hamp}
J.~{Mainprice}, E.~{Akin Sisbot}, L.~{Jaillet}, J.~{Cortés}, R.~{Alami}, and
  T.~{Siméon}, ``Planning human-aware motions using a sampling-based costmap
  planner,'' in \emph{IEEE Int. Conf. Rob. Autom.}, 2011, pp. 5012--5017.

\bibitem{Simeon-T-RRT}
D.~{Devaurs}, T.~{Siméon}, and J.~{Cortés}, ``Enhancing the transition-based
  rrt to deal with complex cost spaces,'' in \emph{2013 IEEE International
  Conference on Robotics and Automation}, 2013, pp. 4120--4125.

\bibitem{Berenson:human-robot-planning}
R.~{Hayne}, R.~{Luo}, and D.~{Berenson}, ``Considering avoidance and
  consistency in motion planning for human-robot manipulation in a shared
  workspace,'' in \emph{IEEE Int. Conf. Rob. Autom.}, 2016, pp. 3948--3954.

\bibitem{Zhao:considering-human-behavior}
X.~{Zhao} and J.~{Pan}, ``Considering human behavior in motion planning for
  smooth human-robot collaboration in close proximity,'' in \emph{2018 27th
  IEEE International Symposium on Robot and Human Interactive Communication
  (RO-MAN)}, 2018, pp. 985--990.

\bibitem{STOMP:2011}
M.~Kalakrishnan, S.~Chitta, E.~Theodorou, P.~Pastor, and S.~Schaal, ``{STOMP}:
  Stochastic trajectory optimization for motion planning,'' in \emph{Proc. IEEE
  ICRA}, Shanghai (China), 2011, pp. 4569--4574.

\bibitem{Tarbouriech:bi-objective}
S.~{Tarbouriech} and W.~{Suleiman}, ``Bi-objective motion planning approach for
  safe motions: Application to a collaborative robot,'' \emph{Journal of
  Intelligent and Robotic Systems}, vol.~99, pp. 45--63, 2020.

\bibitem{Pellegrinelli:POMDP-hindsight-optimization}
S.~Javdani, H.~Admoni, S.~Pellegrinelli, S.~S. Srinivasa, and J.~A. Bagnell,
  ``Shared autonomy via hindsight optimization for teleoperation and teaming,''
  \emph{Int. J. Rob. Res.}, vol.~37, no.~7, pp. 717--742, 2018.

\bibitem{Kanazawa:TRO2019}
A.~{Kanazawa}, J.~{Kinugawa}, and K.~{Kosuge}, ``Adaptive motion planning for a
  collaborative robot based on prediction uncertainty to enhance human safety
  and work efficiency,'' \emph{IEEE Transactions on Robotics}, vol.~35, no.~4,
  pp. 817--832, 2019.

\bibitem{Faroni_CPHS2020}
M.~Beschi, M.~Faroni, C.~Copot, and N.~Pedrocchi, ``How motion planning affects
  human factors in human-robot collaboration,'' in \emph{IFAC Conf.
  Cyber-Physical Human-Systems}, 2020.

\bibitem{Wang:optimal-motion-planning}
Y.~{Wang}, Y.~{Sheng}, J.~{Wang}, and W.~{Zhang}, ``Optimal collision-free
  robot trajectory generation based on time series prediction of human
  motion,'' \emph{IEEE Rob. Autom. Lett.}, vol.~3, no.~1, pp. 226--233, 2018.

\bibitem{pham:topp}
H.~Pham and Q.-C. Pham, ``A new approach to time-optimal path parameterization
  based on reachability analysis,'' \emph{IEEE Transactions on Robotics},
  vol.~34, no.~3, pp. 645--659, 2018.

\bibitem{elbanhawi:sampling-based-review}
M.~Elbanhawi and M.~Simic, ``Sampling-based robot motion planning: A review,''
  \emph{Ieee access}, vol.~2, pp. 56--77, 2014.

\bibitem{Gammel:InformedRRT}
J.~D. Gammell, T.~D. Barfoot, and S.~S. Srinivasa, ``Informed sampling for
  asymptotically optimal path planning,'' \emph{IEEE Transactions on Robotics},
  vol.~34, no.~4, pp. 966--984, 2018.

\bibitem{Gammel:BIT}
J.~Gammell, S.~Srinivasa, and T.~D. Barfoot, ``Batch informed trees (bit*):
  Informed asymptotically optimal anytime search,'' \emph{Int. J. Rob. Res.},
  vol.~39, no.~5, 2020.

\bibitem{Xu:I-FMT}
J.~{Xu}, K.~{Song}, D.~{Zhang}, H.~{Dong}, Y.~{Yan}, and Q.~{Meng}, ``Informed
  anytime fast marching tree for asymptotically optimal motion planning,''
  \emph{IEEE Tran on Industrial Electronics}, vol.~68, pp. 5068--5077, 2021.

\bibitem{Faroni_ETFA2019}
M.~Faroni, M.~Beschi, and N.~Pedrocchi, ``An {MPC} framework for online motion
  planning in human-robot collaborative tasks,'' in \emph{Proc. IEEE Int. Conf.
  on Emerging Tech. and Factory Automation}, 2019.

\bibitem{Kulic:IJRR}
D.~Kulić, C.~Ott, D.~Lee, J.~Ishikawa, and Y.~Nakamura, ``Incremental learning
  of full body motion primitives and their sequencing through human motion
  observation,'' \emph{Int. J. Rob. Res.}, vol.~31, no.~3, pp. 330--345, 2012.

\bibitem{fragkiadaki:RNN}
K.~Fragkiadaki, S.~Levine, P.~Felsen, and J.~Malik, ``Recurrent network models
  for human dynamics,'' in \emph{Proceedings of the IEEE International
  Conference on Computer Vision}, 2015, pp. 4346--4354.

\bibitem{Mainprice:prediction}
P.~{Kratzer}, M.~{Toussaint}, and J.~{Mainprice}, ``Prediction of human
  full-body movements with motion optimization and recurrent neural networks,''
  in \emph{IEEE Int. Conf. Rob. Autom.}, 2020, pp. 1792--1798.

\bibitem{Berret:IOC}
B.~Berret, E.~Chiovetto, F.~Nori, and T.~Pozzo, ``Evidence for composite cost
  functions in arm movement planning: An inverse optimal control approach,''
  \emph{PLOS Computational Biology}, vol.~7, no.~10, pp. 1--18, 2011.

\bibitem{koppula2015anticipating}
H.~S. Koppula and A.~Saxena, ``Anticipating human activities using object
  affordances for reactive robotic response,'' \emph{IEEE Tran on Pattern
  Analysis and Machine Intelligence}, vol.~38, no.~1, pp. 14--29, 2015.

\bibitem{Bajcsy:IJRR}
D.~Fridovich-Keil, A.~Bajcsy, J.~F. Fisac, S.~L. Herbert, S.~Wang, A.~D.
  Dragan, and C.~J. Tomlin, ``Confidence-aware motion prediction for real-time
  collision avoidance1,'' \emph{Int. J. Rob. Res.}, vol.~39, no. 2-3, pp.
  250--265, 2020.

\bibitem{ISOTS15066}
``{ISO/TS 15066:2016 Robots and robotic devices -- Collaborative robots},''
  Int. Org. for Standardization, Geneva, CH, Standard, 2016.

\bibitem{neubauer1989tikhonov}
A.~Neubauer, ``Tikhonov regularisation for non-linear ill-posed problems:
  optimal convergence rates and finite-dimensional approximation,''
  \emph{Inverse problems}, vol.~5, no.~4, p. 541, 1989.

\end{thebibliography}
    
\end{document}